\documentclass[conference]{IEEEtran}
\IEEEoverridecommandlockouts
\usepackage{cite}
\usepackage{graphicx}
\usepackage{cite}
\usepackage{picinpar}
\usepackage{amsmath}
\usepackage{stfloats}
\usepackage{url}
\usepackage{flushend}
\usepackage{colortbl}
\usepackage{soul}
\usepackage{multirow}
\usepackage{pifont}
\usepackage{color}
\usepackage{xcolor}
\usepackage[process=auto]{pstool} % process=all|auto|none
\usepackage{enumerate}
\usepackage{footnote}
\usepackage{graphicx,import}
\usepackage{graphicx}
\usepackage{adjustbox}
\usepackage{float}
\usepackage[caption=false,font=normalsize,labelfont=sf,textfont=sf]{subfig}
\usepackage{alltt}
\usepackage[hidelinks]{hyperref}
\usepackage{enumerate}
\usepackage{siunitx}
\DeclareSIUnit \var { Var }
\usepackage{breakurl}
\usepackage{epstopdf}
\usepackage{pbox}
\usepackage{import}
\usepackage{calc}
\usepackage{algorithm}
\usepackage{algorithmicx}
\usepackage{algpseudocode}
\usepackage{amsthm}
\usepackage{amsmath}
\usepackage{amssymb} 
\usepackage{etoolbox} % For appendix eqn counter from 1.
\usepackage{subcaption} % Required for subfigures
\usepackage{amsmath} % Often useful with figures for captions etc.
\usepackage{lipsum} % For dummy text to show column layout

\usepackage[normalem]{ulem} 
 \usepackage{tabularray}
\usepackage{tabularx,booktabs}

\def\BibTeX{{\rm B\kern-.05em{\sc i\kern-.025em b}\kern-.08em
    T\kern-.1667em\lower.7ex\hbox{E}\kern-.125emX}}
    
\begin{document}

% \title{Resilient Federated LSTM for Electric Vehicle Charging Demand Forecasting Under Cyberattacks\\
\title{Federated Anomaly Detection and Mitigation for EV Charging Forecasting Under Cyberattacks \\

% {\footnotesize \textsuperscript{*}Note: Sub-titles are not captured in Xplore and
% should not be used}
\thanks{This work was partly supported by Innovative Human Resource Development
for Local Intellectualization program through the Institute of IITP grant
funded by the Korean government(MSIT) (IITP-2025-RS-2020-II201612,
50\%) and by Priority Research Centers Program through the NRF funded
by the MEST(2018R1A6A1A03024003, 50\%).}
}

\author{\IEEEauthorblockN{Oluleke Babayomi}
\IEEEauthorblockA{\textit{ICT Convergence Research Center} \\
\textit{Kumoh National Institute of Technology}\\
Gumi, South Korea \\
babayomi@ieee.org}
\and
\IEEEauthorblockN{Dong-Seong Kim }
\IEEEauthorblockA{\textit{IT-Convergence Engineering} \\
\textit{Kumoh National Institute of Technology}\\
Gumi, South Korea \\
dskim@kumoh.ac.kr}
% \and
% \IEEEauthorblockN{3\textsuperscript{rd} Given Name Surname}
% \IEEEauthorblockA{\textit{dept. name of organization (of Aff.)} \\
% \textit{name of organization (of Aff.)}\\
% City, Country \\
% email address or ORCID}
% \and
% \IEEEauthorblockN{4\textsuperscript{th} Given Name Surname}
% \IEEEauthorblockA{\textit{dept. name of organization (of Aff.)} \\
% \textit{name of organization (of Aff.)}\\
% City, Country \\
% email address or ORCID}
% \and
% \IEEEauthorblockN{5\textsuperscript{th} Given Name Surname}
% \IEEEauthorblockA{\textit{dept. name of organization (of Aff.)} \\
% \textit{name of organization (of Aff.)}\\
% City, Country \\
% email address or ORCID}
% \and
% \IEEEauthorblockN{6\textsuperscript{th} Given Name Surname}
% \IEEEauthorblockA{\textit{dept. name of organization (of Aff.)} \\
% \textit{name of organization (of Aff.)}\\
% City, Country \\
% email address or ORCID}
}

\maketitle

\begin{abstract}

Electric Vehicle (EV) charging infrastructure faces escalating cybersecurity threats that can severely compromise operational efficiency and grid stability. Existing forecasting techniques are limited by the lack of combined robust anomaly mitigation solutions and data privacy preservation. Therefore, this paper addresses these challenges by proposing a novel anomaly-resilient federated learning framework that simultaneously preserves data privacy, detects cyber-attacks, and maintains trustworthy demand prediction accuracy under adversarial conditions. The proposed framework integrates three key innovations: LSTM autoencoder-based distributed anomaly detection deployed at each federated client, interpolation-based anomalous data mitigation to preserve temporal continuity, and federated Long Short-Term Memory (LSTM) networks that enable collaborative learning without centralized data aggregation. The framework is validated on real-world EV charging infrastructure datasets combined with real-world DDoS attack datasets, providing robust validation of the proposed approach under realistic threat scenarios. Experimental results demonstrate that the federated approach achieves superior performance compared to centralized models, with 15.2\% improvement in R$^2$ accuracy while maintaining data locality. The integrated cyber-attack detection and mitigation system produces trustworthy datasets that enhance prediction reliability, recovering 47.9\% of attack-induced performance degradation while maintaining exceptional precision (91.3\%) and minimal false positive rates (1.21\%). The proposed architecture enables enhanced EV infrastructure planning, privacy-preserving collaborative forecasting, cybersecurity resilience, and rapid recovery from malicious threats across distributed charging networks.

\end{abstract}

\begin{IEEEkeywords}
Federated learning, LSTM autoencoder, EV charging infrastructure, cybersecurity, anomaly detection
\end{IEEEkeywords}

\section{Introduction}

Electric vehicles (EV) and their charging stations become increasingly interconnected through Internet of Things (IoT) technologies. They form complex cyber-physical systems that integrate seamlessly with power grids, energy management systems, and user interfaces to optimize energy distribution and enhance user experience. However, this interconnectedness introduces critical cybersecurity vulnerabilities that pose significant threats to both operational continuity and public safety. EV charging infrastructure represents a particularly attractive target for malicious actors due to its strategic importance in urban mobility, its direct connection to critical energy grid infrastructure, and its processing of sensitive user data including payment information, location data, and charging patterns. The consequences of successful cyberattacks extend far beyond simple service disruption, potentially triggering cascading failures that compromise grid stability, expose personal data, and create safety hazards for users and surrounding communities. Recent analyses of vulnerabilities in EV charging networks have revealed multiple attack vectors, including weak authentication protocols, inadequate encryption standards, and insufficient network segmentation, highlighting the urgent need for robust cybersecurity frameworks specifically designed for this critical infrastructure.

Distributed Denial-of-Service (DDoS) attacks represent one of the most pervasive and damaging forms of cyberattacks targeting modern infrastructure systems \cite{Zainudin2023}. These attacks involve malicious attempts to overwhelm target systems by flooding them with traffic from multiple compromised sources, effectively rendering services unavailable to legitimate users. In the context of EV charging stations, DDoS attacks manifest through coordinated efforts to overload communication networks, control systems, and data processing capabilities, creating severe operational disruptions that cascade throughout the charging ecosystem. DDoS attacks can destabilize power grid operations
by causing sudden load drops or unexpected spikes, potentially
triggering protective relay operations and widespread power
outages.

Current research in EV charging demand forecasting encompasses a diverse range of methodological approaches, from traditional statistical models to advanced deep learning architectures. Various models including autoregressive integrated moving average (ARIMA), support vector machines, random forests, and deep neural networks have been applied to predict charging demand patterns. However, most existing approaches focus primarily on accuracy optimization under ideal conditions, with limited consideration of operational resilience under adversarial scenarios.

Federated Learning (FL) has emerged as a promising paradigm for time series forecasting in industrial IoT contexts, offering significant advantages in privacy preservation, distributed training capabilities, and system scalability. \cite{Saputra2019} proposes Energy Demand Prediction with Federated Learning for Electric Vehicle Networks, utilizing traditional neural networks for distributed demand forecasting. However, this approach suffers from limited capability to capture long-term temporal dependencies inherent in energy consumption patterns, while lacking mechanisms to detect and mitigate anomalies in EV data or network communications.

Long Short-Term Memory (LSTM) networks have demonstrated superior performance for time series prediction tasks, particularly excelling due to their gating mechanisms that effectively solve the vanishing gradient problem, enabling them to learn long-term temporal dependencies that traditional neural networks and other machine learning methods cannot capture effectively \cite{Shanmuganathan2022}. Ref. \cite{Shrestha2024} presents a framework for identifying anomalies in industrial data gathered from remote terminal devices deployed at substations in smart electric grid systems, employing LSTM and autoencoders with Mean Standard Deviation (MSD) and Median Absolute Deviation (MAD) approaches for anomaly detection while utilizing FL to preserve data privacy. \cite{Douaidi2023} demonstrated the effectiveness of Federated Deep Learning with LSTM for predicting EV station occupancy, achieving 86.21\% accuracy and 91.49\% F1-score while guaranteeing privacy and minimizing data transfer costs.

Recent cybersecurity-focused research has begun addressing DDoS attack detection in EV infrastructure. Authors in \cite{Talaat2025} evaluate several machine learning models, including Random Forest, Gradient Boosting Machine, K-Nearest Neighbors, and Multilayer Perceptron for model interpretability and detection accuracy of DDoS attacks in EV charging stations using Personalized Federated Learning.

Despite these advances, critical limitations persist in current state-of-the-art approaches. Most forecasting models assume clean, anomaly-free data inputs, rendering them unrealistic and unreliable in cyberattack environments where data integrity is compromised. 
Based on the preceding literature review, several critical research gaps emerge that this study aims to address. There exists a notable lack of robust forecasting solutions for EV charging demand that explicitly account for and actively mitigate the impact of cyberattacks like DDoS at the edge computing level. Current approaches lack effective, privacy-preserving anomaly detection mechanisms that are seamlessly integrated into distributed forecasting pipelines, creating vulnerabilities in real-world deployment scenarios.

% The primary objective of this study is to solve the combined challenge of resilient charging demand prediction while maintaining strict data privacy requirements, developing a comprehensive framework that bridges the gap between cybersecurity and demand forecasting in federated learning environments.

This research makes several significant contributions to the field of resilient IoT systems for smart city infrastructure. First, we propose a federated LSTM forecasting system that enables collaborative learning across multiple charging stations while maintaining data locality and privacy, ensuring that sensitive operational data never leaves individual station boundaries. Second, we develop a comprehensive cybersecurity framework incorporating an integrated LSTM autoencoder-based anomaly detection system and a sophisticated interpolation-based anomaly mitigation mechanism that preserves temporal continuity while restoring data integrity. Third, we provide comprehensive empirical validation through extensive experiments using real-world EV charging infrastructure datasets combined with real-world DDoS attack datasets, providing robust validation of the proposed approach under realistic threat scenarios. These contributions extend our previous work in \cite{Babayomi2025a} to distributed clients requiring privacy guarantees.
The rest of the paper is organized as follows: Sec. \ref{Sec:Methodology} discusses the proposed methodology, Sec. \ref{Sec:DiscussionResults} discusses the experimental results, and Sec. \ref{Sec:Conclusion} concludes the paper.

\begin{figure}
    \centering
    \includegraphics[width=0.9\linewidth]{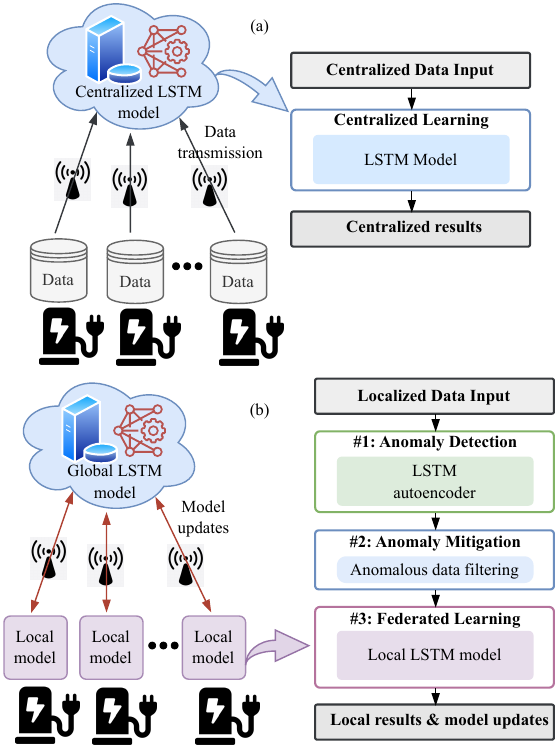}
    \caption{EV charging demand forecasting. (a) Conventional centralized learning approach. (b) Proposed trustworthy anomaly-resilient federated LSTM framework.}
    \label{Fig:Conv_and_proposed_Resilient_FL_LSTM}
\end{figure}

% \begin{figure*}
%     \centering
%     \includegraphics[width=0.99\linewidth]{Timeseries_FL_centralized.pdf}
%     \caption{Time series}
%     \label{fig:Timeseries_FL_centralized}
% \end{figure*}

\section{Methodology}
\label{Sec:Methodology}
The proposed technique is shown in Fig. \ref{Fig:Conv_and_proposed_Resilient_FL_LSTM}, and the following discussion will provide details of its operation.

\subsection{ Data Description and Preparation}

This study analyzes Shenzhen's electric vehicle charging patterns using comprehensive data collected from September 2022 to February 2023. The dataset encompasses real-time charging station information including availability, pricing, and locations, originally collected at 5-minute intervals via a mobile platform and subsequently organized into a 1-hour resolution region-level dataset. Weather data from meteorological observatories were also included as contextual information, though not directly incorporated into the forecasting models presented in this work.

Shenzhen serves as an ideal case study due to its pioneering role in large-scale vehicle electrification infrastructure development. The city's advanced EV ecosystem provides valuable insights for other municipalities developing similar infrastructure and seeking to understand urban charging patterns. From the comprehensive dataset containing 331 total traffic zones, this research focuses on three specific traffic zones: '102', '105', and '108', named Clients 1, 2 and 3, respectively. Each contains 4,344 timestamps corresponding to the study period.

The data preparation process followed standard machine learning practices. A temporal train-test split strategy was implemented with 80\% of the data allocated for training and 20\% for testing. MinMaxScaler normalization was applied independently to each client's raw data across all experimental scenarios (before and after anomaly injection/filtering) to ensure consistent scaling ranges between 0 and 1. For LSTM model input preparation, a sequence length of 24 timestamps was employed, corresponding to 24 hours of historical look-back given the 1-hour data resolution, enabling the models to capture daily charging patterns and dependencies.

\subsection{Anomaly Injection and Detection Framework}
To evaluate system resilience against cyberattacks, simulated DDoS attack patterns were implemented based on actual network traffic characteristics documented in the literature. The attack simulation parameters were derived from real-world measurements where normal IP traffic averaged 33,000 packets per second (p/s) while attack traffic reached 350,500 p/s, representing a 10.6 times intensity multiplier over normal conditions with 100ms time slots.

These network-level characteristics were systematically translated into DDoS-like anomalies for EV charging volume data by applying intensity multipliers derived from the documented attack patterns. The anomalies manifested as irregular volume spikes that disrupted normal charging demand patterns, simulating the effect of coordinated cyberattacks on charging infrastructure data integrity.

The core anomaly detection mechanism was implemented through the EVChargingAnomalyFilter class, featuring an LSTM Autoencoder architecture for unsupervised anomaly detection. The autoencoder employed an encoder-decoder structure with LSTM layers (50→25 neurons in encoder, 25→50 neurons in decoder) and incorporated dropout regularization (0.2) to prevent overfitting. The system was trained exclusively on normal (non-anomalous) data segments to establish baseline reconstruction patterns.
Anomaly identification utilized reconstruction error analysis, specifically mean squared error (MSE) between input sequences and their reconstructed counterparts. The 98th percentile threshold was applied to MSE values computed on the training set, establishing the anomaly detection boundary. Points exceeding this threshold were flagged as anomalous.
The $filter\_anomalies$ method implemented anomaly mitigation through sophisticated linear interpolation. The algorithm identified consecutive anomalous segments, allowing for small gaps ($\leq2$ timestamps) to maintain continuity, and applied interpolation between non-anomalous boundary points. This approach preserved data temporal structure while removing attack-induced distortions.

Three distinct data scenarios were generated for each client to enable comprehensive performance analysis: Clean Data (original, unmodified charging patterns), Attacked Data (incorporating DDoS-like anomalies), and Filtered Data (attacks detected and mitigated through interpolation).

\subsection{Forecasting Model Architecture}
Two primary forecasting architectures were implemented to evaluate the effectiveness of federated versus centralized approaches:

\subsubsection{Centralized LSTM Model} The centralized architecture employed a Sequential model with LSTM (50) followed by Dense (10, activation='relu') and final Dense (1) output layers. Input data consisted of reshaped combined sequences from all clients, processed jointly on anomalous data without preprocessing. This approach simulated traditional centralized systems operating on potentially compromised data streams.

\subsubsection{Federated LSTM Model} The federated architecture utilized identical local client models with the same LSTM structure (LSTM units=50, Dense layers configuration) but trained independently on local datasets. The Federated Averaging mechanism facilitated global model coordination through weight synchronization while preserving data locality. Only model parameters were exchanged between clients, maintaining privacy and data sovereignty principles essential for industrial applications.
Key hyperparameters were standardized across experiments: $SEQUENCE\_LENGTH=24$ (hours), $LSTM\_UNITS=50$, $EPOCHS\_PER\_ROUND=10$, $FEDERATED_ROUNDS=5$, $LEARNING\_RATE=0.001$, and $batch size=32$. Early stopping with patience$=10$ was implemented to prevent overfitting during autoencoder training.

\begin{figure}
    \centering
    \includegraphics[width=0.75\linewidth]{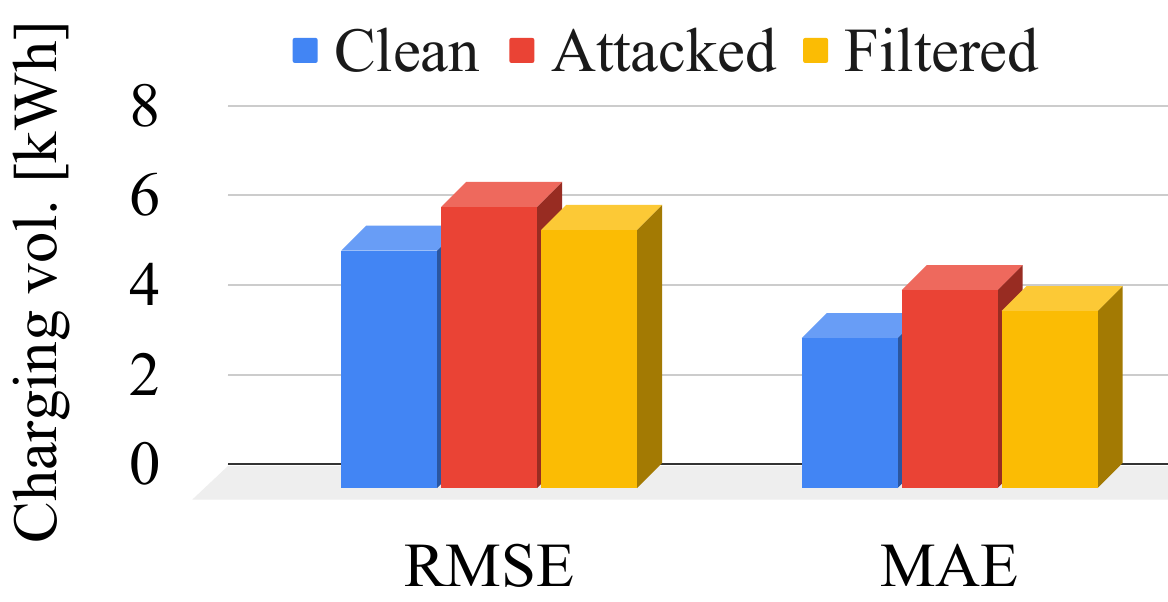}
    \caption{Performance of the proposed anomaly-resilient federated LSTM for Client 1.}
    \label{Fig:rmse_mae_FL_plot}
\end{figure}

\begin{figure}
    \centering
    \includegraphics[width=0.75\linewidth]{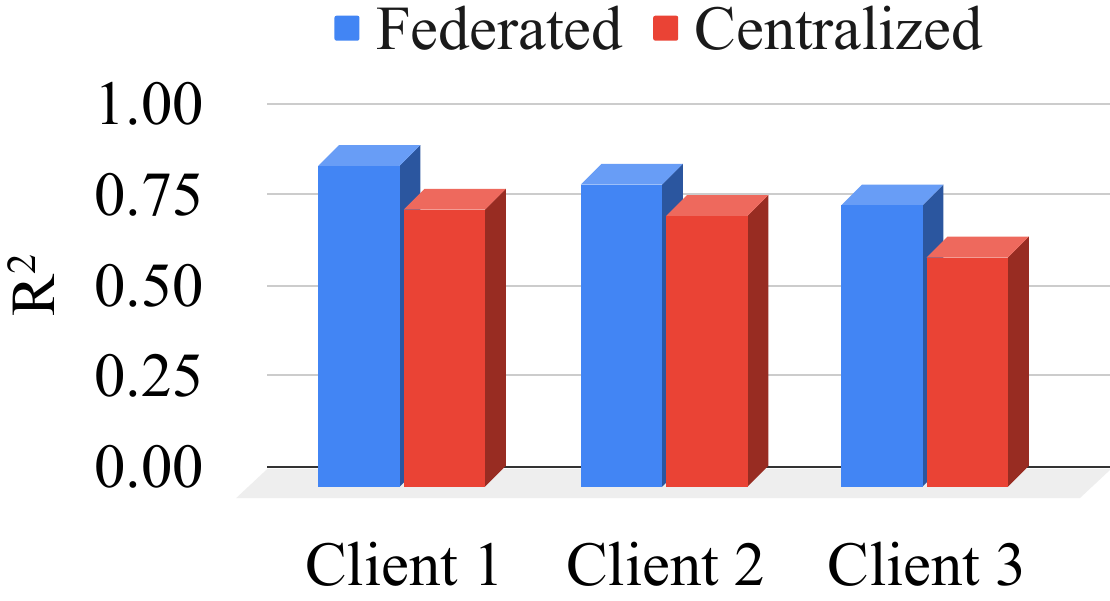}
    \caption{Comparison of R$^2$ for both federated LSTM and centralized LSTM for filtered data.}
    \label{fig:Federated_n_Centralized_R2_plot}
\end{figure}

\section{Discussion of Results}
\label{Sec:DiscussionResults}

\subsection{Experimental Design and Evaluation Framework}
The experimental framework comprised four distinct scenarios designed to isolate and quantify the impact of cyberattacks and mitigation strategies:
\begin{enumerate}
    \item Federated LSTM (Clean Data): Baseline performance on uncompromised data. 
\item Federated LSTM (Attacked Data): Performance degradation under simulated DDoS attacks. 
\item Federated LSTM (Filtered Data): Recovery effectiveness with anomaly detection and filtering.
\item Centralized LSTM (Filtered Data): Traditional centralized approach operating on the same filtered data as the federated approach.
\end{enumerate}

It is important to note that the centralized versus federated comparison (scenarios 3 and 4) utilizes identical filtered datasets across all clients. This ensures a fair architectural comparison where both approaches benefit from the same anomaly detection and filtering pipeline, isolating the performance differences attributable to the learning architecture itself rather than data preprocessing advantages.

Performance evaluation employed standard regression metrics: Mean Absolute Error (MAE), Root Mean Squared Error (RMSE), and coefficient of determination (R$^2$). These metrics provided comprehensive assessment of prediction accuracy and model reliability across different attack scenarios.
Anomaly detection effectiveness was quantified through classification metrics including Overall Detection Precision, Recall, F1-Score, True Attacks Detected ratio, and False Positive Rate. Per-client metrics were computed for traffic zones 102, 105, and 108 to analyze spatial heterogeneity in detection performance. The discussion is based on the performances of Client 1, but Tables provide a broader picture for all three Clients.

%Statistical significance was assessed using T-statistics and P-values to validate performance differences between experimental scenarios. For consistent visualization and comparison, primary results focused on Client 1 (zone 102) test data, supplemented by comprehensive multi-client analysis for centralized model evaluation.

 \subsection{Impact of Simulated DDoS Attacks on Forecasting Performance}
The experimental results reveal a more nuanced impact of DDoS attacks on federated LSTM forecasting performance than initially anticipated. Considering Client 1 (see Table \ref{Tab:Complete_comparison_clean}), the R$^2$ coefficient showed a moderate decline from 0.9075 (Clean Data) to 0.8707 (Attacked Data), representing a 4.0\% reduction in explained variance. This degradation was accompanied by proportional increases in error metrics: MAE increased from 3.3859 to 4.4134 (30.3\% increase) and RMSE rose from 5.3162 to 6.2835 (21.7\% increase).

While these performance reductions are statistically significant, they indicate a level of inherent resilience in the federated LSTM architecture against the simulated DDoS attacks. The relatively modest impact suggests that distributed learning systems may possess some natural robustness to certain types of cyberattacks, particularly when the attacks manifest as volume spikes that can be partially absorbed by the model's capacity to handle normal demand variations. However, the attack simulation, based on adapted documented real-world network data to the present study, still demonstrates measurable threats to smart grid and EV infrastructure forecasting systems that warrant proactive mitigation strategies.

% Table 2: Complete Performance Comparison (Clean Version)
\begin{table}[!t]
\centering
\caption{Complete performance comparison for Client 1.}
\label{Tab:Complete_comparison_clean}
\begin{tabular}{llcccc}
\toprule
\textbf{Scenario} & \textbf{Architecture} & \textbf{MAE} & \textbf{RMSE} & \textbf{R$^2$} & \textbf{Time (s)} \\
\midrule
Clean Data & Federated & 3.3859 & 5.3162 & 0.9075 & 80.85 \\
Attacked Data & Federated & 4.4134 & 6.2835 & 0.8707 & 80.33 \\
Filtered Data & Federated & 3.9801 & 5.7921 & 0.8883 & 85.95 \\
Filtered Data & Centralized & 6.1644 & 8.6040 & 0.7536 & 101.46 \\
\bottomrule
\end{tabular}
\end{table}

\begin{table}[h]
\centering
\caption{Client-Specific Anomaly Detection Results}
\label{tab:client_anomaly_results}
\begin{tabular}{lccc}
\toprule
\textbf{Client} & \textbf{Precision} & \textbf{Recall} & \textbf{F1} \\
\midrule
1 (102) & 0.907 & 0.584 & 0.710 \\
2 (105) & 0.955 & 0.591 & 0.730 \\
3 (108) & 0.859 & 0.354 & 0.501 \\
\bottomrule
\end{tabular}
\end{table}

\subsection{Effectiveness of Anomaly Detection and Filtering Mechanisms}

The implemented LSTM autoencoder-based anomaly detection system demonstrated meaningful recovery capabilities. The Federated LSTM performance on Filtered Data achieved an R$^2$ of 0.8883, representing a 47.9\% recovery of the attack-induced performance loss, as shown in Fig. \ref{Fig:rmse_mae_FL_plot}. While not achieving complete restoration to baseline performance, this recovery validates the effectiveness of unsupervised anomaly detection in mitigating cyberattack effects.

Analysis of detection metrics reveals a precision-focused detection strategy. The high overall precision (0.913) and low false positive rate (1.21\%) indicate that flagged anomalies are highly likely to be genuine attacks, minimizing unnecessary data alteration. 

Per-client analysis (see Table \ref{tab:client_anomaly_results}) revealed significant spatial heterogeneity in detection performance: Client 1 (zone 102) achieved exceptional precision (0.907) with moderate recall (0.584, F1=0.710); Client 2 (zone 105) demonstrated the highest precision (0.955) with similar recall (0.591, F1=0.730); while Client 3 (zone 108) showed good precision (0.859) but notably lower recall (0.354, F1=0.501). This variation suggests that zone 108's charging patterns may be more difficult to distinguish from attack signatures, potentially due to naturally occurring demand spikes that resemble attack patterns.

% Table 1: Client-Specific Performance Comparison (Clean Version)
\begin{table}[!t]`
\centering
\caption{Client-specific performance comparison for filtered data.}
\label{tab:client_comparison_clean}
\begin{tabular}{llccc}
\toprule
\textbf{Client (Zone)} & \textbf{Architecture} & \textbf{MAE} & \textbf{RMSE} & \textbf{R$^2$} \\
\midrule
\multirow{2}{*}{\textbf{Client 1 (102)}} & Federated & 3.9801 & 5.7921 & 0.8883 \\
& Centralized & 6.8277 & 8.4567 & 0.7646 \\
\midrule
\multirow{2}{*}{\textbf{Client 2 (105)}} & Federated & 5.2215 & 5.5876 & 0.8350 \\
& Centralized & 6.5100 & 8.1582 & 0.7463 \\
\midrule
\multirow{2}{*}{\textbf{Client 3 (108)}} & Federated & 5.0459 & 6.2328 & 0.7792 \\
& Centralized & 5.1554 & 9.1659 & 0.6356 \\
\bottomrule
\end{tabular}
\end{table}

\subsection{Federated versus Centralized Architecture Performance}
A substantial architectural performance advantage emerges when comparing the federated approach for Client 1 (R$^2$ =0.8883) against the centralized approach (R$^2$ =0.7536), both operating on identical filtered datasets (see Fig. \ref{fig:Federated_n_Centralized_R2_plot}). The federated model significantly outperformed the centralized model by 15.2\% in R$^2$, while achieving 54.9\% lower MAE (3.9801 vs. 6.1644) and 48.5\% lower RMSE (5.7921 vs. 8.6040).
This substantial performance superiority is particularly significant because both approaches utilized the same cleaned, filtered data, eliminating data quality as a confounding factor. The performance difference is therefore attributable to fundamental architectural advantages of the federated learning paradigm for spatially heterogeneous time-series forecasting. Table \ref{tab:client_comparison_clean} shows data for the three Clients.

\subsection{Local Specialization versus Global Generalization} The federated architecture enables each local model to specialize in zone-specific charging patterns for each of zone 102, 105 and 108. In contrast, the centralized model must identify a compromise solution that performs adequately across all zones but may not excel at capturing any specific local pattern.

\subsubsection{Training Dynamics and Optimization} Federated local models can reach superior local optima by focusing on consistent, zone-specific patterns within their 4,344 timestamp datasets. The centralized approach, while training on a larger combined dataset (13,032 timestamps), must simultaneously optimize across heterogeneous patterns from different urban contexts, potentially leading to suboptimal parameter configurations that represent compromises between conflicting pattern requirements.

The centralized model's highly heterogeneous per-client performance (Client 1: R²=0.7646, Client 2: R$^2=0.7463$, Client 3: R$^2=0.6356$) demonstrates this compromise effect, where the single global model performs inconsistently across different spatial contexts. Notably, the centralized model's worst performance occurs at Client 3 (zone 108), which also showed the most challenging anomaly detection characteristics, suggesting that zone 108’s charging patterns are inherently more complex and benefit significantly from specialized local modeling.

\subsubsection{Computational Efficiency} The federated approach also demonstrated superior computational efficiency, completing training in 85.95 seconds compared to 101.46 seconds for centralized training, an 18.1\% reduction in training time. This efficiency gain reflects the distributed computational load and parallel processing capabilities inherent in federated architectures.

\subsubsection{Architectural Implications for Industrial IoT} This finding demonstrates that federated learning architectures provide substantial advantages over centralized approaches for spatially distributed industrial systems, independent of privacy or security considerations. The 15.2\% performance improvement represents a significant gain that would translate to meaningful operational improvements in real-world EV charging network management and grid stability predictions.

\subsection{Computational Efficiency and Practical Deployment Considerations}
The federated training approach demonstrated superior computational efficiency, completing in 85.95 seconds compared to 101.46 seconds for centralized training, an 18.1\% reduction in training time. This efficiency advantage stems from distributed computational load, parallel processing capabilities, and the inherent efficiency of optimizing smaller, more homogeneous local datasets rather than large, heterogeneous centralized aggregations.
The training time consistency across federated scenarios (80.8s for clean data, 80.3s for attacked data, 85.9s for filtered data) indicates stable computational requirements regardless of data quality, suggesting robust scalability for real-world deployments. This consistency is particularly valuable for industrial applications requiring predictable computational resource allocation.
From a practical deployment perspective, the federated approach offers multiple operational advantages beyond performance and efficiency. The distributed architecture provides inherent system resilience through redundancy, eliminates single points of failure, and enables continued operation even when individual nodes experience downtime. The 15.2\% performance advantage combined with 18.1\% computational efficiency improvement creates a compelling business case for federated implementations in large-scale EV charging networks.

\subsection{Implications for Industrial IoT and Smart Grid Architecture}
The results reveal significant implications beyond cybersecurity for industrial IoT systems, particularly regarding the fundamental choice between centralized and distributed learning architectures. The demonstrated superiority of federated learning over centralized approaches--even when operating on identical, clean data--challenges traditional assumptions about the necessity of data centralization for optimal machine learning performance.

\subsubsection{Architectural Paradigm Shift} For spatially distributed industrial systems such as smart grids, EV charging networks, and distributed manufacturing, the findings suggest that federated architectures can deliver superior performance by leveraging local specialization while maintaining collaborative intelligence. This represents a paradigm shift from viewing federated learning primarily as a privacy-preserving compromise to recognizing it as potentially superior architecture for heterogeneous industrial environments.

\subsubsection{Scalability and Resilience} The federated approach's computational efficiency advantage (18.1\% faster training) combined with superior prediction accuracy creates a compelling case for large-scale industrial deployments. The distributed architecture inherently provides system resilience through redundancy while enabling local optimization that centralized systems cannot achieve.
The successful integration of anomaly detection within the federated framework demonstrates the feasibility of implementing comprehensive security measures without sacrificing performance or requiring data centralization. This capability is particularly relevant for critical infrastructure where both security and operational excellence are mandatory requirements.

\subsubsection{Limitations and Future Research Directions}
Several limitations should be acknowledged in this work. First, the study relies on real-world DDoS attack dataset which was adapted to the present study. While this approach enables controlled experimentation, validation with actual cyberattack data on EV infrastructure would strengthen the findings. Second, the anomaly detection focuses specifically on sustained high-volume irregular spikes; other attack vectors such as subtle data manipulation or temporal pattern disruption warrant investigation.
The linear interpolation-based filtering method, while effective, represents a basic mitigation approach. More sophisticated reconstruction techniques leveraging deep generative models or advanced time-series imputation methods could potentially improve recovery performance while reducing false positive impacts.
Future research directions include: (1) validation with real-world cyberattack datasets on EV charging infrastructure, (2) investigation of additional attack vectors including false data injection and sophisticated adversarial patterns, and (3) exploration of advanced filtering and reconstruction techniques beyond linear interpolation,.
Furthermore, the integration of additional contextual data (weather patterns, traffic conditions, economic factors) could enhance both baseline forecasting accuracy and anomaly detection precision. Multi-modal attack detection combining network-level and application-level anomaly indicators represents another promising research avenue.

 \section{Conclusion}
 \label{Sec:Conclusion}
This research presents a novel federated LSTM framework that successfully addresses the dual challenges of cybersecurity and demand forecasting accuracy in EV charging infrastructure. The proposed system demonstrates superior performance over centralized approaches, achieving higher prediction accuracy while providing 18.1\% faster training and maintaining data privacy. The integrated anomaly detection mechanism effectively identifies DDoS attacks with 91.3\% precision and 1.21\% false positive rate, recovering 47.9\% of attack-induced performance degradation through sophisticated interpolation-based mitigation. These findings challenge traditional assumptions about centralized learning necessity, revealing federated architectures as potentially superior solutions for heterogeneous industrial environments. The work establishes a paradigm shift toward distributed intelligence in critical infrastructure, demonstrating that privacy-preserving approaches can simultaneously enhance both security and operational performance. Future research should explore additional attack vectors and advanced reconstruction techniques to further strengthen cyber-resilience capabilities.

%----BIBLIOGRAPHY
\bibliographystyle{IEEEtran}
\bibliography{Ref_Aug2025}

\end{document}